\documentclass[letterpaper, 10 pt, conference]{ieeeconf}  

\IEEEoverridecommandlockouts                              

\usepackage{graphicx} 
\usepackage{amsmath} 
\usepackage{amssymb}  
\usepackage{subcaption}
\usepackage{multirow}
\usepackage[table]{xcolor}
\usepackage{tikz}
\usepackage{tikz-3dplot}
\usepackage{placeins}
\usetikzlibrary {arrows.meta} 
\usetikzlibrary{calc,backgrounds}
\usepackage{threeparttable}
\usepackage[bookmarks=false]{hyperref}

\title{\LARGE \bf
ISOPoT: Imaging Sonar Odometry by Point Tracking
}

\author{Jaša Samec$^{1}$, Vid Rijavec$^{1}$, Marko Peljhan$^{3,4}$, Aleksander Grm$^{2}$, \\ Andrej Androjna$^{2}$, Danijel Skočaj$^{1}$, Matej Dobrevski$^{1}$
 \thanks{*This work was supported by the European Defense Agency, project SABUVIS II.}
 \thanks{$^{1}$Jaša Samec, Vid Rijavec, Danijel Skočaj and Matej Dobrevski are with the Faculty of Computer and Information Science, University of Ljubljana,
         Večna pot 113, 1000 Ljubljana, Slovenia.
         {\tt\small matej.dobrevski@fri.uni-lj.si}}%
 \thanks{$^{2}$Aleksander Grm and Andrej Androjna are with the Faculty of Maritime Studies and Transport, University of Ljubljana,
         Pot pomorščakov 4, 6320 Portorož, Slovenia.}%
 \thanks{$^{3}$Marko Peljhan is with the OZON Research Group, TIMTEC,
         Goriska cesta 6c, Vipava, 5271, Slovenia.}%
 \thanks{$^{4}$Marko Peljhan is also with the MAT Systemics, University of California, Santa Barbara, 
         3309 Phelps Hall, Santa Barbara, 93106-6065, California, United States of America.}%
 }

\begin{document}

\maketitle
\thispagestyle{empty}
\pagestyle{empty}

\begin{abstract}

Reliable navigation in underwater environments remains a key challenge in marine robotics. In such scenarios, forward-looking sonars are a natural choice for long-range perception, offering wide coverage even in turbid, low-visibility conditions.  However, sonar images are inherently noisy, contain artifacts, and lack rich semantic structure, causing standard computer vision methods for keypoint detection and matching to perform poorly. In this paper, we introduce ISOPoT, an imaging sonar odometry method based on modern point tracking techniques. We propose a sonar odometry pipeline that uses multi-frame point tracks as its primary correspondence representation, augmented with lightweight optimizations to improve robustness. We evaluated the proposed method on the Aracati 2017 dataset, as well as on an internal sonar dataset collected in real-world underwater environments. Our results show that ISOPoT outperforms previous state-of-the-art methods consistently in both sonar-only scenarios and in multi-sensor settings.

\end{abstract}

\section{INTRODUCTION}

Underwater environments remain challenging for mobile robot navigation. Water significantly attenuates much of the electromagnetic spectrum, limiting the use of GNSS and other radio-based systems. Optical cameras are useful at short range in clear water, but in many practical settings, turbidity limits visibility to only a few meters (or less). Consequently, acoustic sensing is often the only exteroceptive modality that provides reliable perception at ranges relevant for navigation, and forward-looking imaging sonars have become common sensors for underwater vehicles. In this work, we consider the problem of estimating planar $SE(2)$ motion from forward-looking sonar imagery.

Imaging sonars produce two-dimensional acoustic intensity images whose statistics differ sharply from those of optical images. Sonar frames exhibit speckle noise, sensor artifacts, weak local texture, and strong appearance changes induced by viewpoint and insonification angle. As a result, standard vision pipelines based on detecting, describing, and matching local keypoints are unreliable on sonar data. The problem is further compounded by the scarcity of high-quality labeled underwater datasets, since accurate ground truth is difficult and expensive to obtain.

Existing imaging sonar odometry methods~\cite{sonic,diso} still inherit the same core weakness: they formulate correspondence as pairwise keypoint matching. This formulation can be brittle on sonar imagery. Local patches are often ambiguous, and independently matched correspondences provide little protection against globally inconsistent motion. The result is noisy estimation and brittle failure in low-structure or artifact-heavy regions. Many systems compensate by incorporating auxiliary odometry to initialize motion or constrain the search space, but this does not resolve the underlying mismatch between sonar imagery and the pairwise matching paradigm.

\begin{figure}[!t]
  \centering
    \includegraphics[width=\linewidth]{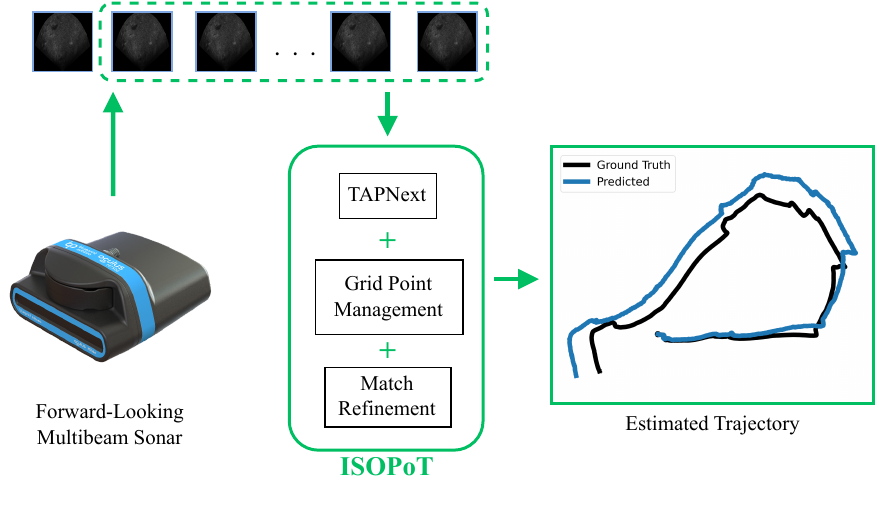}
  \caption{Overview of the proposed ISOPoT pipeline. Given a sliding window of forward-looking sonar images, the method tracks points across the window and estimates the robot's planar $SE(2)$ motion.}
  \label{fig:MethodDiagram}
\end{figure}

We take a different approach. The ambiguity of local patches suggests that an approach that considers the images globally might be more suitable for estimating corresponding regions. Robustness can be further improved by leveraging temporal consistency across multiple frames. Recent tracking-any-point methods such as AllTracker~\cite{alltacker2025}, CoTracker3~\cite{cotracker3}, and TAPNext~\cite{tapnext2025} are designed precisely to exploit both the global information across the image as well as the temporal structure of the features. Although trained on optical videos, we find that several of these models transfer effectively to forward-looking sonar imagery.

Based on this observation, we propose \emph{Imaging Sonar Odometry by Point Tracking} (ISOPoT), a sonar-only pipeline for planar motion estimation from a sliding window of forward-looking sonar images. ISOPoT uses TAPNext as a tracking backbone, a grid-based point manager to preserve spatial coverage, and a lightweight correlation-based refinement stage that converts tracked points into a single motion estimate under a static-scene assumption. By replacing brittle local matching with motion-consistent multi-frame tracking, ISOPoT yields substantially more stable correspondences in cluttered, weakly textured, and artifact-heavy sonar sequences. Figure~\ref{fig:MethodDiagram} summarizes the proposed pipeline.

The main contributions of this paper are threefold. First, we show that modern tracking-any-point models transfer effectively to forward-looking sonar imagery, and that multi-frame point tracking is a more suitable correspondence primitive for sonar odometry than pairwise keypoint matching. Second, we introduce ISOPoT, which combines TAPNext-based tracking, grid-based point management, and lightweight global refinement for robust planar sonar odometry. Third, we evaluate the method on the public Aracati 2017 dataset~\cite{aracati2017} and an in-house dataset, and show that it achieves significantly lower odometry error than prior sonar baselines on the evaluated sequences, in both sonar-only and auxiliary-sensor assisted settings.

\section{Background and Related Work}
\label{sec:rw}

\subsection{Sonar Image Formation and Motion Model}

\begin{figure}[!t]
  \centering
    \includegraphics[width=0.99\linewidth]{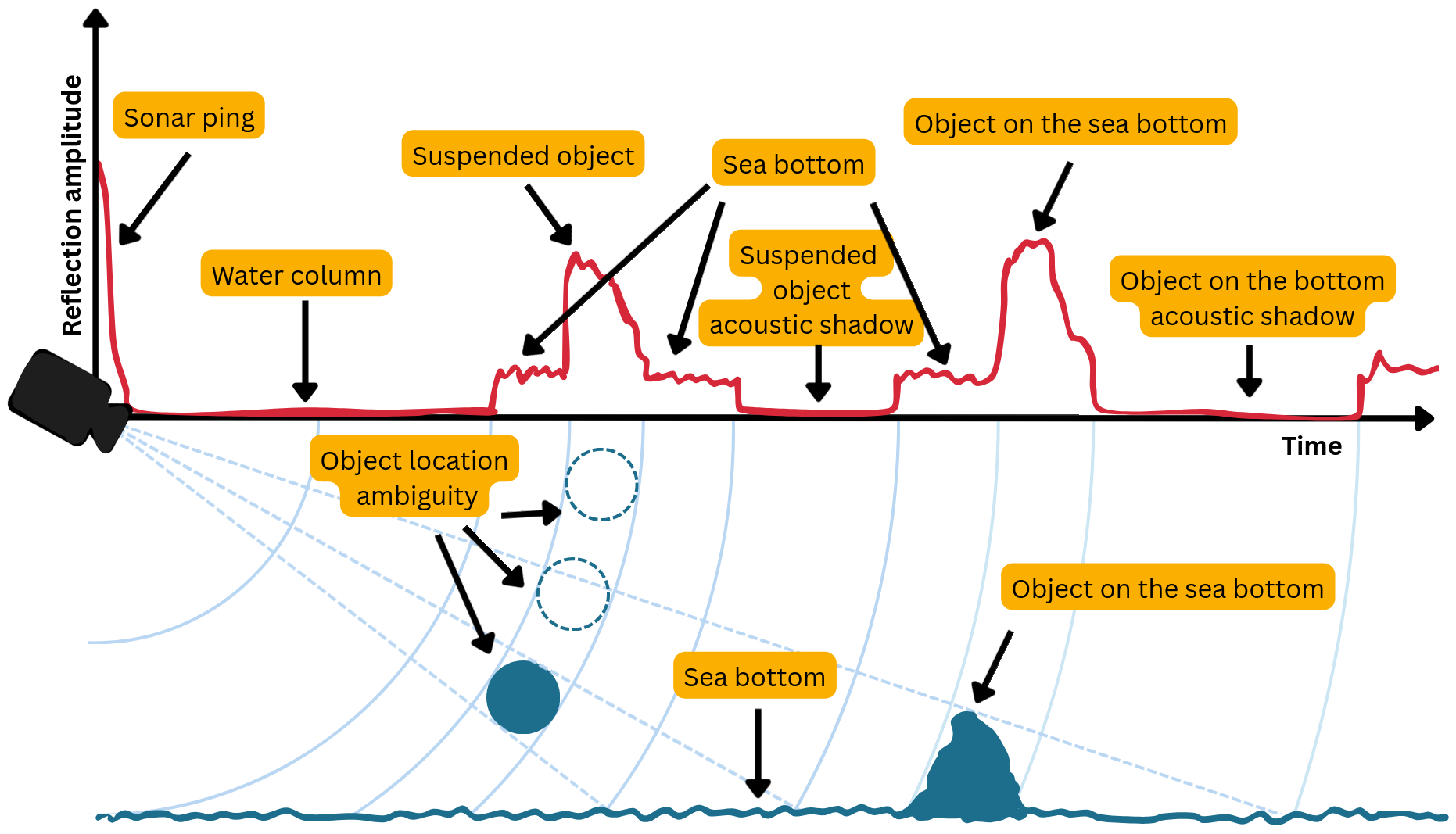}
  \caption{Image formation for a single sonar beam. For a fixed beam direction, the returned intensity at a given time-of-flight aggregates reflections from an iso-range arc within the beam aperture. A sonar pixel therefore determines range and bearing in the sensor frame, but not elevation.}
  \label{fig:SonarSingleBeam}
\end{figure}

We focus on forward-looking multibeam sonar, which forms a two-dimensional acoustic intensity image by transmitting a fan of beams and recording returned energy as a function of time-of-flight and beam direction. Figure~\ref{fig:SonarSingleBeam} illustrates the geometry for a single beam. After calibration, each image pixel $\mathbf{p}$ is associated with a range--bearing pair $(r(\mathbf{p}), \theta(\mathbf{p}))$ in the sonar frame. Unlike a perspective camera, however, a sonar pixel does not correspond to a unique 3-D point: elevation within the beam aperture is unobserved, and the measured intensity may combine echoes from multiple reflectors at the same range and bearing.

Under the standard assumptions of forward-looking sonar odometry, a predominantly static scene, limited pitch and roll variation, and approximately planar vehicle motion, the observable component of motion can be modeled in the horizontal sensor plane. Defining the Cartesian representation of a scene point as
\begin{equation}
\mathbf{x}(\mathbf{p})=
\begin{bmatrix}
r(\mathbf{p})\cos\theta(\mathbf{p})\\
r(\mathbf{p})\sin\theta(\mathbf{p})
\end{bmatrix},
\end{equation}
where a scene point observed in consecutive frames satisfies
\begin{align}
\mathbf{x}_{k+1} &\approx \mathbf{R}(\Delta\psi_k)\mathbf{x}_k+\mathbf{t}_k, \\
T_{k+1\leftarrow k} &=
\begin{bmatrix}
\mathbf{R}(\Delta\psi_k) & \mathbf{t}_k\\
\mathbf{0}^\top & 1
\end{bmatrix}\in SE(2),
\end{align}
where $\Delta\psi_k$ and $\mathbf{t}_k$ are the relative yaw and planar translation between frames $k$ and $k+1$. This relation is inherently approximate, as elevation is not observed. Compounding this, sonar-specific phenomena, namely speckle, acoustic shadows, and multipath, introduce a stochastic visibility process that causes scene points to appear or disappear between frames independently of the inter-frame motion. The key estimation problem is therefore to recover a set of correspondences jointly consistent with a single dominant planar motion, rather than to rely on isolated local matches.

\subsection{Imaging Sonar Odometry and SLAM}

Existing imaging sonar odometry and SLAM methods can be broadly grouped by the front-end used to establish inter-frame correspondences. Geometry-based approaches align larger structures extracted from sonar images, for example by applying ICP to feature sets or local maps~\cite{bluerov}. Other methods use local keypoints and descriptors, either handcrafted or learned. The method in~\cite{pose_odometry} extracts CNN features at detected keypoints and searches for matches within regions constrained by an initial odometry estimate, which is later refined in a pose graph. DISO~\cite{diso} likewise depends on external odometry for coarse alignment and then refines the solution by minimizing intensity differences between matched keypoints, followed by factor-graph optimization. SONIC~\cite{sonic} removes the dependence on external motion priors and performs sonar-only matching using CNN features together with a coarse-to-fine correlation strategy.

Despite these differences, most prior methods share the same front-end assumption: motion is recovered from pairwise local correspondences between consecutive frames. Global consistency is typically imposed only afterward, through RANSAC, pose-graph optimization, factor graphs, or auxiliary odometry. This design is often brittle on sonar data, where local appearance is weakly distinctive and strongly affected by noise, incidence angle, and view-dependent acoustic shadows~\cite{FourierSonar}. Back-end optimization can improve global trajectory consistency, but it does not resolve failures in the front-end correspondence stage. This observation motivates our shift from pairwise local matching to multi-frame point tracking.

\subsection{Point Tracking for Sonar Correspondence}

Tracking Any Point (TAP) methods estimate the trajectories of arbitrary query points across a video sequence. Unlike pairwise matching, TAP models exploit temporal context, visibility reasoning, and iterative refinement to maintain correspondences over long temporal windows. Early work such as TAP-Net and the TAP-Vid benchmark formalized the task of predicting point coordinates and visibility across frames~\cite{tapvid}. Subsequent methods improved robustness through iterative and multi-frame inference. PIPs models each track as a persistent particle and refines tracks recurrently over time~\cite{pip}, Context-PIPs incorporates additional spatial context~\cite{contextpips}, and TAPIR combines per-frame matching with iterative local refinement~\cite{tapir}. Joint tracking methods such as CoTracker and CoTracker3 further couple many trajectories within a single model, allowing information sharing across points and improving robustness to occlusion and out-of-view motion~\cite{cotracker3,cotracker}. Recent approaches such as AllTracker and TAPNext push this trend further by framing tracking as general long-range correspondence estimation or sequence prediction~\cite{alltacker2025,tapnext2025}.

This evolution is particularly relevant for sonar. In contrast to RGB video, sonar imagery violates appearance-constancy assumptions more severely: local texture is sparse, repeated seabed patterns are common, and intensity varies with range, viewpoint, and insonification angle~\cite{FourierSonar}. As a result, independently matched local patches are prone to ambiguity and track swapping. The temporal evolution of many points, however, still contains a strong geometric signal. Multi-frame point tracking is therefore a more natural correspondence primitive for sonar odometry than independent pairwise keypoint matching.

Among the trackers we evaluated, TAPNext~\cite{tapnext2025} provided the most reliable tracks on sonar sequences. TAPNext formulates point tracking as sequential prediction of trajectory tokens conditioned on past frames and query points, enabling causal and online inference with explicit visibility estimates. These properties are well aligned with our setting: the model can operate on a sliding temporal window, its visibility output helps reject unreliable tracks, and its multi-frame inference reduces the brittleness of pairwise local matching. For these reasons, we adopt TAPNext as the tracking backbone of the proposed ISOPoT pipeline and build the remainder of the method around robust motion estimation from tracked points.

\section{Method}
\label{sec:method}

\begin{figure}[!t]
  \centering
    \includegraphics[width=0.95\linewidth]{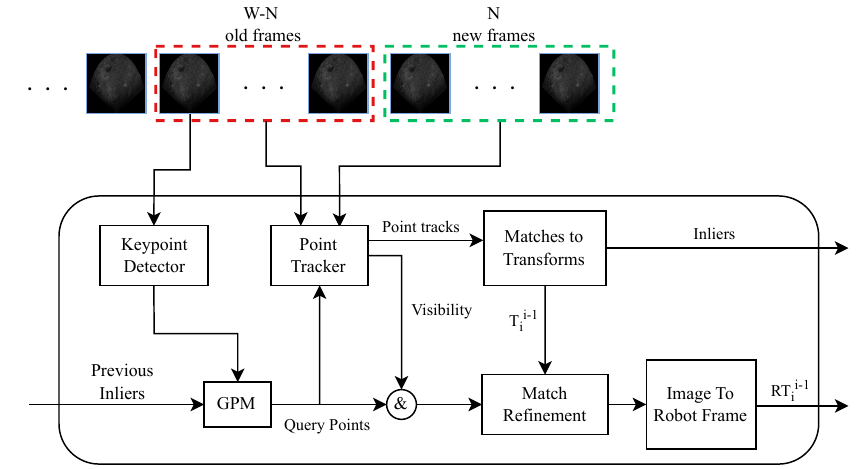}
  \caption{Overview of ISOPoT. A sliding window containing \(B\) overlapping frames and \(N\) new frames is processed at each iteration. Sobel points detected in the first frame are merged with surviving points from the previous window by the Grid Point Manager. TAPNext tracks the selected queries across the window, RANSAC estimates coarse image-plane transforms, and the Match Refinement module produces the final motion estimates. Only the pose increments associated with the newly arrived frames are appended to the output trajectory.}
  \label{fig:MethodPartsDiagram}
\end{figure}

Figure~\ref{fig:MethodPartsDiagram} summarizes the proposed pipeline. At each iteration, ISOPoT processes a sliding window
\begin{equation}
\mathcal{W}=\{F_1,\dots,F_W\}
\end{equation}
of sonar images, where the first \(B=W-N\) frames overlap with the previous window and the remaining \(N\) frames are newly arrived observations. In all experiments, \(W\) is set equal to the sonar frame rate in frames 
per second, meaning one window spans approximately one second of data, and we set \(B=3\). This overlap acts as a warm-up period for TAPNext: on sonar imagery, the earliest predictions are often less reliable than later predictions, and the overlap allows us to suppress these transient errors.

\subsection{ISOPoT}

ISOPoT is a two-stage correspondence pipeline. TAPNext provides temporally consistent but coarse point tracks across the window, and a lightweight refinement stage converts these tracks into a single globally consistent motion estimate. The method assumes a predominantly static scene and the planar-motion model introduced in Section~\ref{sec:rw}.

\paragraph{Query point initialization.}
Let \(Q=\{\mathbf{q}_j\}_{j=1}^M\), with \(\mathbf{q}_j \in \mathbb{R}^2\), denote the query points placed on the first frame \(F_1\) of the current window. The query set is formed by combining two sources:  
(i) new points detected on \(F_1\) using a Sobel-based detector, and  
(ii) points that survived as reliable inliers in the overlapping portion of the previous window.

The Sobel detector computes image gradients, applies non-maximum suppression, and retains points whose gradient magnitude lies above the 98th percentile. In our experiments, these Sobel points were more stable on sonar imagery than AKAZE~\cite{akaze} or SuperPoint~\cite{superpoint}: AKAZE often produced too few points, whereas lowering the SuperPoint threshold to recover more meaningful sonar structure also introduced many unstable responses.

The combined point set is filtered by the Grid Point Manager (Section~\ref{sec:GPM}), which enforces spatial coverage and outputs the final query set for TAPNext. Given \(Q\) and the window \(\mathcal{W}\), TAPNext predicts for each point \(\mathbf{q}_j\) its trajectory
\begin{equation}
\{\mathbf{q}_{j,t}\}_{t=1}^W
\end{equation}
and corresponding visibility flags
\begin{equation}
\{v_{j,t}\}_{t=1}^W.
\end{equation}

\paragraph{Coarse motion estimation from tracked points.}
For each target frame \(F_t\), \(t\in\{2,\dots,W\}\), we estimate a 2-D Euclidean transform in image coordinates from the tracked points in \(F_1\) to their predicted locations in \(F_t\). Using homogeneous coordinates \(\bar{\mathbf{q}}_j=[q_{x,j},q_{y,j},1]^\top\) and the projection operator \(\pi([x,y,1]^\top)=[x,y]^\top\), we compute the coarse transform
\begin{equation}
\tilde{T}^{\mathrm{img}}_{t\leftarrow 1}
=
\arg\min_{T \in SE(2)}
\sum_{j \in \mathcal{V}_t}
\left\|
\mathbf{q}_{j,t} - \pi\!\left(T \bar{\mathbf{q}}_j\right)
\right\|_2^2,
\label{eq:coarse_transform}
\end{equation}
where \(\mathcal{V}_t\) contains points that TAPNext predicts as visible in frame \(t\). The transform is estimated robustly with RANSAC using a 3-pixel inlier threshold, which also yields an inlier/outlier partition of the tracked points.

Rather than fitting motion only between consecutive frames, we estimate transforms to a common reference frame \(F_1\). On sonar sequences, this is more robust because displacement relative to the first frame is larger than between adjacent frames, making drifting or stationary points easier to identify. Consecutive image-plane increments are then recovered as
\begin{equation}
\tilde{T}^{\mathrm{img}}_{t+1\leftarrow t}
=
\tilde{T}^{\mathrm{img}}_{t+1\leftarrow 1}
\left(\tilde{T}^{\mathrm{img}}_{t\leftarrow 1}\right)^{-1},
\qquad t=1,\dots,W-1.
\label{eq:relative_from_absolute}
\end{equation}
These coarse increments are subsequently refined.

\subsubsection{Grid Point Manager}
\label{sec:GPM}

The Grid Point Manager (GPM) maintains a set of reliable query points while ensuring that they remain well distributed throughout the image. This is important because RANSAC is less stable when most points are concentrated in a small image region.

We divide the image into a \(16\times16\) grid and allow at most five query points per cell. The GPM merges newly detected Sobel points with points carried over from the previous iteration, then enforces the per-cell limit. When a cell contains more than five candidates, excess points are removed at random. This prevents over-populated regions from dominating the motion estimate while giving newly detected points a chance to replace stale ones.

A point is removed from the maintained set if it satisfies any of the following conditions:  
(i) it is classified as an outlier during RANSAC-based motion estimation in the overlapping portion of the window,  
(ii) TAPNext predicts it as not visible, or  
(iii) it falls outside the valid sonar scan region.  
The output of the GPM is the final query set passed to TAPNext in the current iteration.

\subsubsection{Match Refinement}
\label{sec:MR}

TAPNext provides accurate coarse tracks, but it predicts each point independently and does not directly enforce a single global motion hypothesis. Small per-point errors therefore remain, even when the overall motion direction is correct. We address this with a lightweight global refinement stage that uses local feature correlation to optimize one transform per target frame.

\begin{figure}[!t]
  \centering
    \includegraphics[width=0.95\linewidth]{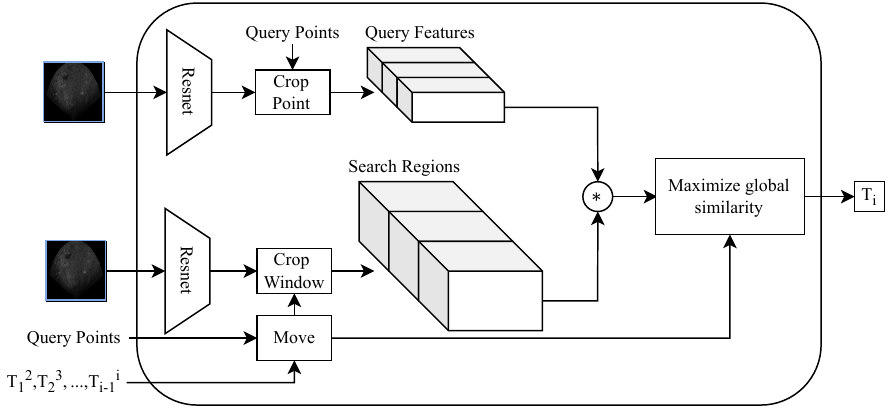}
  \caption{Match refinement. Query points from the first frame are propagated to the target frame using the coarse transform. Dense ResNet features are correlated within local search regions, producing one heatmap per point. A global optimizer then refines a single \(SE(2)\) transform by maximizing the summed correlation across all reliable points.}
  \label{fig:MatcherDiagram}
\end{figure}

The refinement module, illustrated in Figure~\ref{fig:MatcherDiagram}, takes as input the coarse transforms \(\tilde{T}^{\mathrm{img}}_{t\leftarrow 1}\), the query points on \(F_1\), and the corresponding image pairs \((F_1,F_t)\). We first propagate each query point to frame \(F_t\) using the coarse transform. The reference image \(F_1\) and target image \(F_t\) are then passed through the first layer of a ResNet50 backbone~\cite{resnet} pretrained on ImageNet1K~\cite{IMAGENET}. For each query point, we extract a reference descriptor from \(F_1\) and a larger search region around its propagated location in \(F_t\). Correlating the reference descriptor with the search region yields a heatmap \(H_{j,t}\) that scores candidate locations for point \(j\) in frame \(t\).

The final refined transform is obtained by maximizing the summed correlation over all reliable points:
\begin{equation}
\hat{T}^{\mathrm{img}}_{t\leftarrow 1}
=
\arg\max_{T \in SE(2)}
\sum_{j \in \mathcal{R}_t}
H_{j,t}\!\left(\pi\!\left(T \bar{\mathbf{q}}_j\right)\right),
\label{eq:refine_transform}
\end{equation}
where \(\mathcal{R}_t \subseteq \mathcal{V}_t\) contains only points that are predicted visible by TAPNext and classified as inliers by the coarse RANSAC fit. We solve \eqref{eq:refine_transform} by gradient ascent. Optimizing a single global objective reduces the influence of noisy local matches while enforcing consistency with the underlying planar-motion model.

After refinement, consecutive increments are recomputed from the common reference frame as
\begin{equation}
\hat{T}^{\mathrm{img}}_{t+1\leftarrow t}
=
\hat{T}^{\mathrm{img}}_{t+1\leftarrow 1}
\left(\hat{T}^{\mathrm{img}}_{t\leftarrow 1}\right)^{-1}.
\label{eq:refined_relative}
\end{equation}
These refined image-plane transforms are then converted to planar robot-frame pose increments according to the sonar geometry described in Section~\ref{sec:rw}. Only the increments associated with the \(N\) newly arrived frames are appended to the output trajectory; the overlapping frames are used solely to stabilize tracking and refinement.

\subsubsection{Optional Auxiliary-Sensor Correction}

The core ISOPoT pipeline is sonar-only. However, to enable a fair comparison with methods that additionally use motion sensors, we also evaluate an optional correction stage that combines ISOPoT with external odometry and, when available, magnetometer heading. This stage is not part of the learned model; it is included only to test whether low-confidence sonar estimates can be detected and corrected when auxiliary information is available.

Let \(M_t^{\mathrm{in}}\) be the number of RANSAC inliers in frame \(t\) and \(M_t^{\mathrm{trk}}\) the number of tracked points considered for motion estimation. We define the inlier ratio
\begin{equation}
\rho_t = \frac{M_t^{\mathrm{in}}}{M_t^{\mathrm{trk}}},
\end{equation}
and convert it to a confidence weight using
\begin{equation}
w_t = \sigma\!\left(s(\rho_t-c)\right),
\label{eq:conf_weight}
\end{equation}
where \(\sigma(\cdot)\) is the sigmoid and \(s\), \(c\) are two fixed scalars setting its slope and offset.

Let \((\hat{\mathbf{t}}^{\mathrm{iso}}_t,\hat{\psi}^{\mathrm{iso}}_t)\) denote the planar translation and yaw predicted by ISOPoT for timestep \(t\), and let \((\hat{\mathbf{t}}^{\mathrm{odo}}_t,\hat{\psi}^{\mathrm{odo}}_t)\) denote the corresponding estimate from an external odometry source. We interpolate translation as
\begin{equation}
\hat{\mathbf{t}}_t
=
w_t \hat{\mathbf{t}}^{\mathrm{iso}}_t
+
(1-w_t)\hat{\mathbf{t}}^{\mathrm{odo}}_t.
\label{eq:trans_fusion}
\end{equation}
In our experiments, yaw drift is the dominant long-term failure mode when only sonar is used. Therefore, when a bounded-error heading source such as a magnetometer is available, we simply replace the predicted yaw with the measured heading,
\begin{equation}
\hat{\psi}_t = \hat{\psi}^{\mathrm{mag}}_t.
\label{eq:yaw_replace}
\end{equation}
This simple correction proved more reliable than blending two drifting orientation estimates.

\section{Evaluation}
\label{sec:evaluation}

We evaluate ISOPoT on two datasets with different characteristics: the public Aracati 2017 dataset~\cite{aracati2017} and a new dataset collected for this work, Portoroz 2025. The evaluation has three goals. First, we compare ISOPoT against prior sonar odometry baselines in both sonar-only and auxiliary-sensor-assisted settings. Second, we analyze how the behavior of the methods changes across two qualitatively different sonar domains. Third, we study the effect of the tracking backbone and of the main components of the proposed pipeline.

\subsection{Aracati 2017}

\begin{figure}[]
  \centering
    \includegraphics[width=\linewidth]{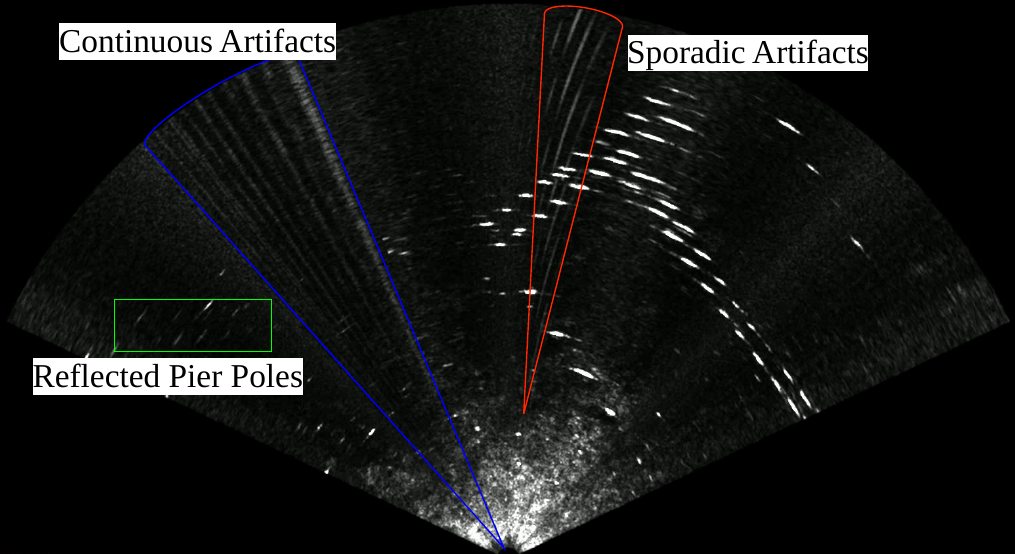}
  \caption{Typical failure modes in the Aracati 2017 dataset. Blue regions indicate persistent stripe artifacts, the green region shows reflections of pier poles moving opposite to the true structures, and the red region marks occasional transient artifacts with comparatively small impact on performance.}
  \label{fig:AracatiIssues}
\end{figure}

This dataset is particularly challenging because the sonar imagery contains several recurring artifacts, shown in Figure~\ref{fig:AracatiIssues}. The most prominent issue is the presence of large stripe artifacts that persist over long portions of the sequence and make affected image regions unreliable for correspondence estimation. In addition, the dataset contains reflection artifacts that resemble real structures, for example mirrored copies of pier poles that move in the opposite direction from the physical scene. Such effects violate the dominant planar-motion assumption used by many front-end estimators. Finally, large overexposed regions suppress subtle seabed structure that would otherwise provide useful motion cues in open-water segments.

\subsection{Portoroz 2025}

Portoroz 2025 was collected in the marina of the Faculty of Maritime Studies and Transport, University of Ljubljana, using the RAS-HA-X25 autonomous vehicle developed within the SABUVIS II project and equipped with an Oculus 750d forward-looking sonar. The dataset consists of two sequences, denoted \emph{Boot} and \emph{Lawnmower}, containing 1452 and 4080 images, respectively, each with spatial resolution \(1220 \times 610\).

\begin{figure}[!t]
  \centering
    \includegraphics[width=\linewidth]{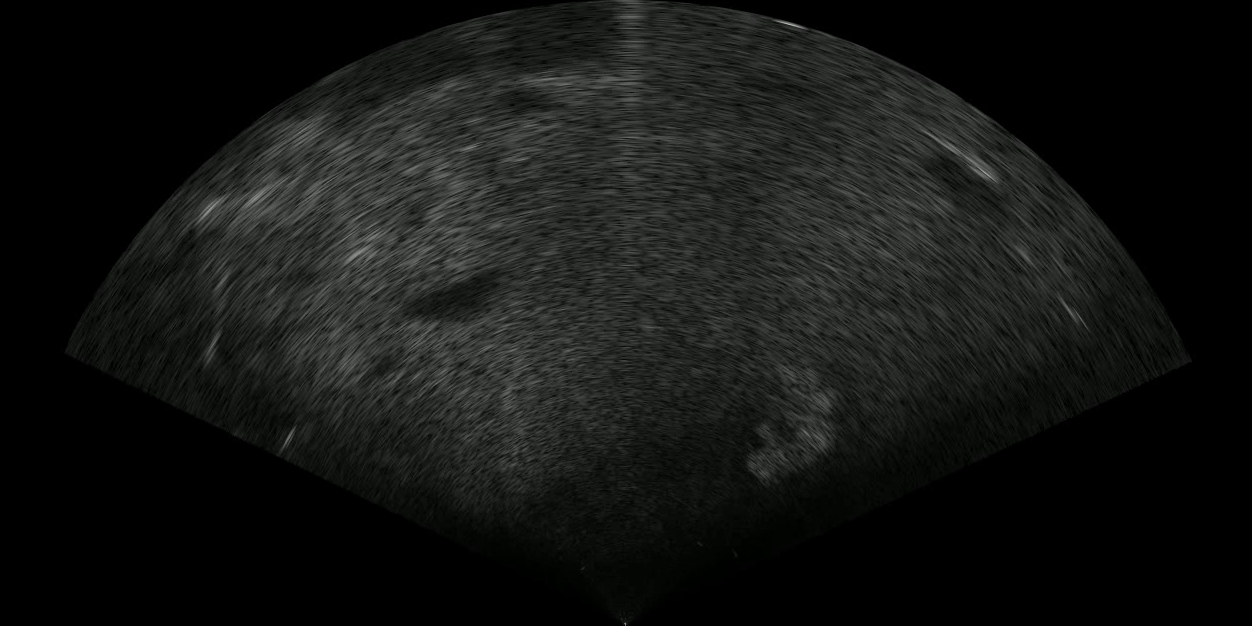}
    \caption{Example sonar image from the Portoroz 2025 dataset. Compared with Aracati 2017, the scene contains weaker object-centric landmarks and more subtle seabed structure, which is difficult to capture with local keypoint detectors but still informative for motion estimation.}
  \label{fig:portoroz}
\end{figure}

Compared with Aracati 2017, Portoroz 2025 contains fewer isolated, high-contrast landmarks and more seabed-dominated regions in which the useful motion signal is distributed across weak but spatially extended acoustic structure. This makes the dataset particularly challenging for methods that rely on detecting and matching a small number of distinctive local keypoints. An example frame is shown in Figure~\ref{fig:portoroz}.

\subsection{Dataset comparison}

Although both datasets contain forward-looking sonar imagery, they represent different correspondence regimes. Aracati 2017 contains some clearly distinguishable structures, such as pier poles and shoreline boundaries, but it is also heavily affected by persistent artifacts and reflections. Portoroz 2025, in contrast, contains fewer strong discrete landmarks and instead relies more on subtle distributed structure in seabed-dominated scenes.

A simple illustration of this difference is given by the AKAZE detector. On Aracati 2017, AKAZE returns between 248 and 2933 keypoints per frame. On Portoroz 2025, the maximum number of detected keypoints is 342, and in some frames no keypoints are detected at all. Importantly, these low keypoint counts do not imply that the image is uninformative for odometry. Rather, they indicate that the available information is poorly represented by sparse local detectors. This difference is one of the main motivations for using multi-frame point tracking instead of pairwise keypoint matching.

\subsection{Experimental setup}

For Aracati 2017, we follow the evaluation protocol used in DISO and use the same evaluation toolbox~\cite{eval_tool}. The full sequence is divided into three sections (S1, S2, S3), as in prior work. For Portoroz 2025, we evaluate on the two available sequences, \emph{Boot} and \emph{Lawnmower}.

For each sequence, we report Absolute Trajectory Error (ATE), aligned only to the first pose, and Relative Error (RE). ATE measures long-horizon drift and therefore reflects how the method would behave in a deployment setting, but it is sensitive to a small number of large failures. RE is computed on trajectory segments of length equal to \(10\%\) of the sequence and is less sensitive to isolated outliers, making it more informative about local odometric accuracy. On Aracati 2017 the heading sensor is mounted above the waterline and thus provides near ground-truth heading, so any method that directly adopts this heading attains near-zero rotation error; the meaningful comparison in the assisted setting is therefore on translation and ATE.

\subsection{Results}

\subsubsection{Aracati 2017}

Table~\ref{tab:AracatiCombined} summarizes the quantitative results on Aracati 2017. In the sonar-only setting, ISOPoT substantially outperforms SONIC in all three sections and by a large margin in both ATE and RE. This confirms that multi-frame point tracking is significantly more reliable than pairwise keypoint matching on this artifact-heavy dataset.

The comparison with DISO must be interpreted separately, since DISO relies on auxiliary odometry and magnetometer measurements to operate. In the odometry+magnetometer-assisted setting, ISOPoT again achieves the best results. In particular, it yields the lowest ATE in all three sections and the lowest relative translation error overall. These results indicate that improving the sonar front-end remains beneficial even when auxiliary motion information is available.

\begin{figure*}[!t]
 \centering
    \begin{subfigure}{0.33\linewidth}
      \centering
      \includegraphics[width=\linewidth]{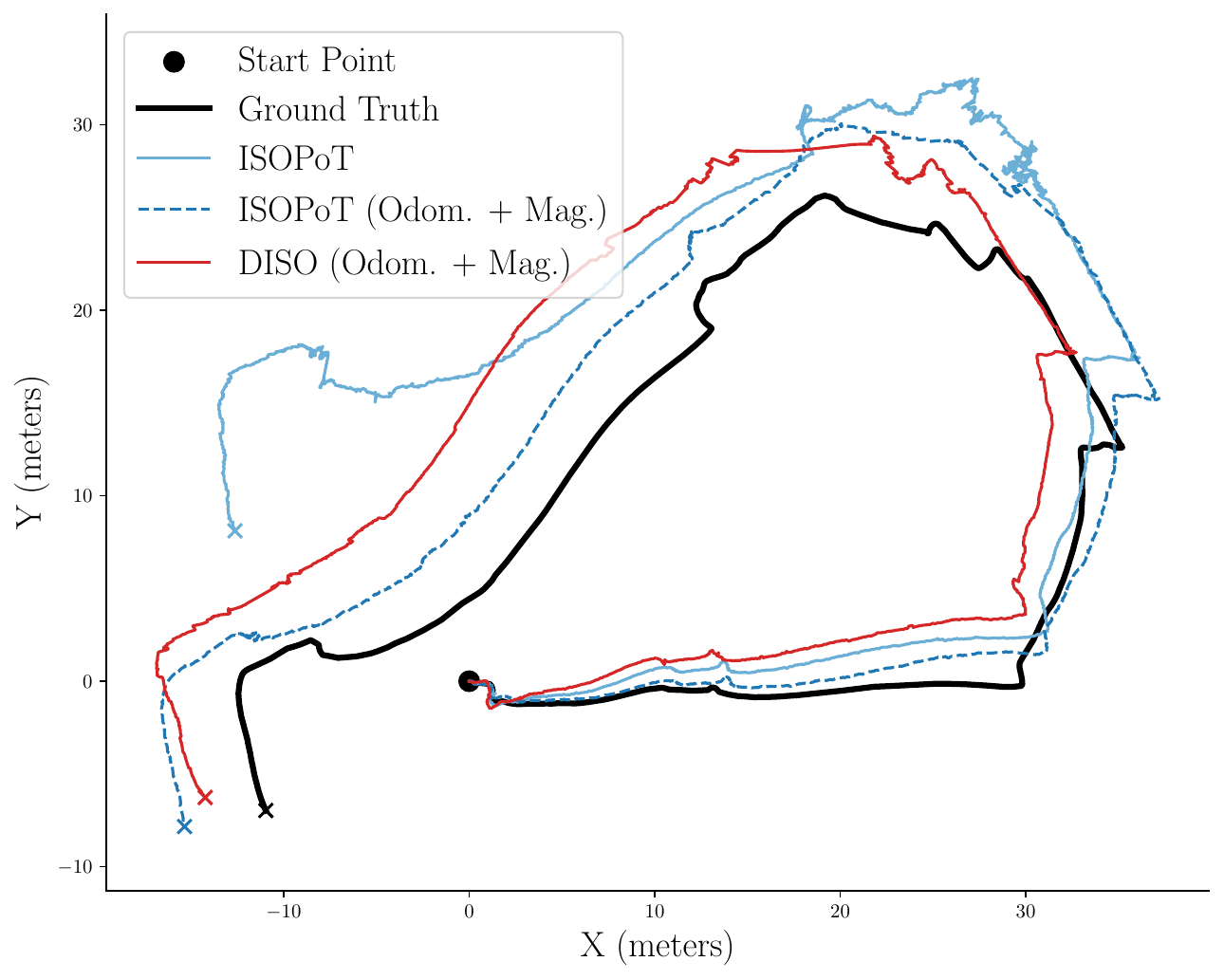}
      \caption{Section 1}
    \end{subfigure}\hspace{0em}%
    \begin{subfigure}{0.33\linewidth}
      \centering
      \includegraphics[width=\linewidth]{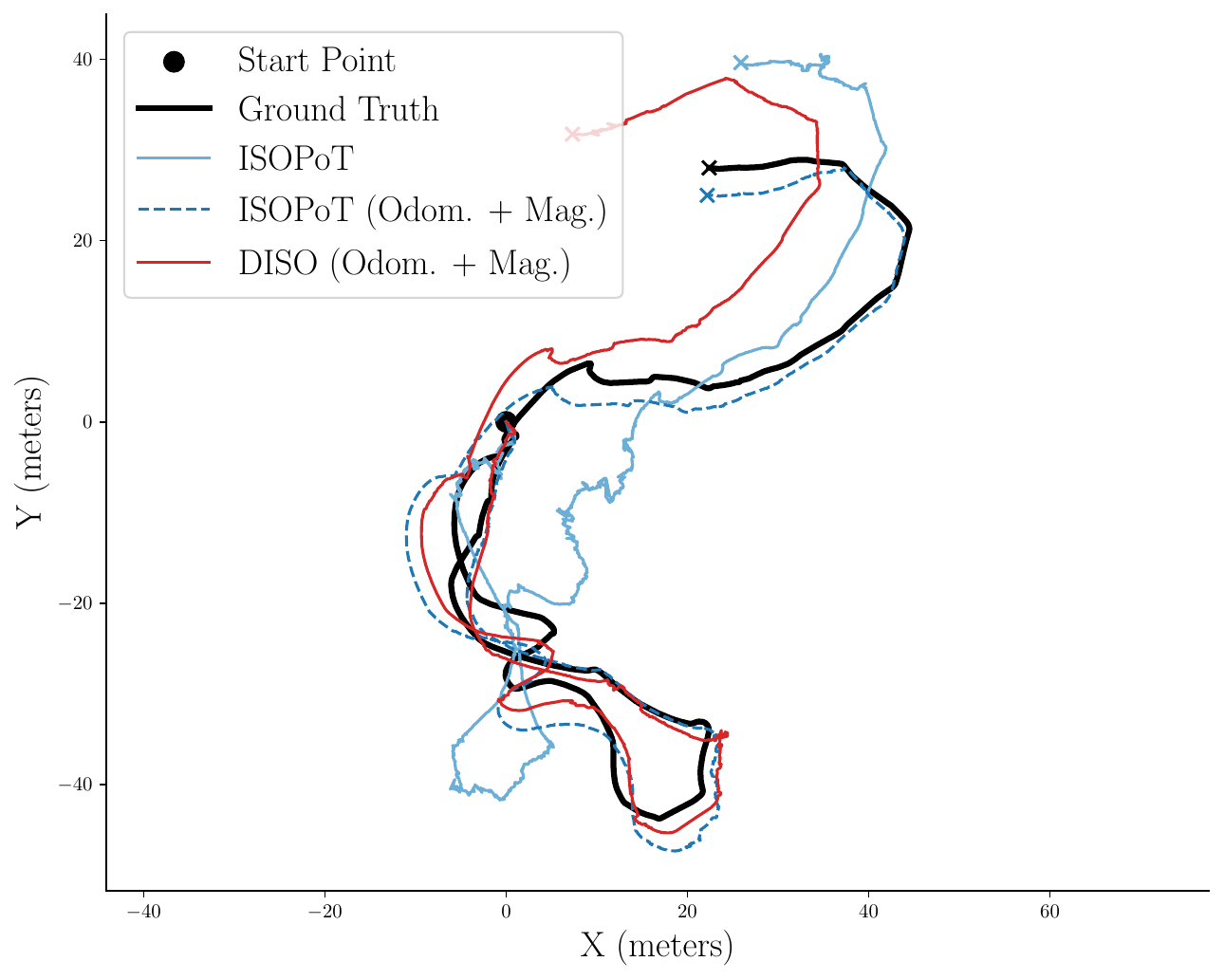}
      \caption{Section 2}
    \end{subfigure}\hspace{0em}%
    \begin{subfigure}{0.33\linewidth}
      \centering
      \includegraphics[width=\linewidth]{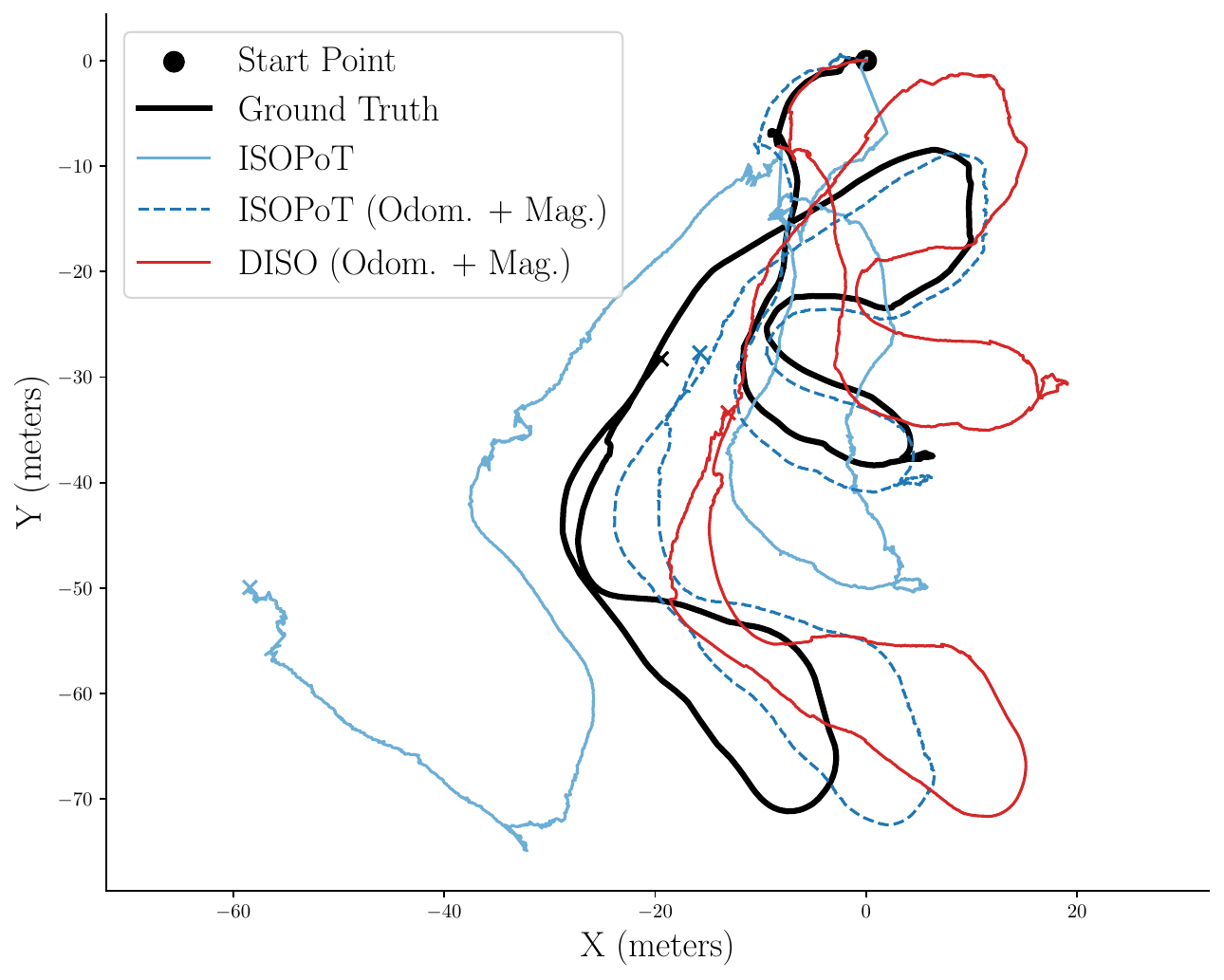}
      \caption{Section 3}
    \end{subfigure}
 \caption{Qualitative comparison on Aracati 2017. ISOPoT without correction follows the overall trajectory but accumulates drift in visually weak segments. Adding odometry correction substantially stabilizes the estimate. SONIC is omitted for readability because its predictions deviate strongly and would clutter the plots.}
 \label{fig:AracatiSections}
\end{figure*}

The main remaining weakness of ISOPoT on Aracati appears in the final part of the trajectory, where a small number of poor predictions cause a visible increase in ATE. This behavior is illustrated in Figure~\ref{fig:AracatiSections}. Because ATE is sensitive to such failures, the relative error in Table~\ref{tab:AracatiCombined} provides a clearer view of local performance. Under this metric, the advantage of ISOPoT over DISO becomes even more apparent.

\begin{table}[!t]
\centering
\small
\setlength{\tabcolsep}{2.5pt}
\begin{threeparttable}
\begin{tabular}{llccccc}
\hline
\multirow{2}{*}{Aux. info.} & \multirow{2}{*}{Method} & \multicolumn{3}{c}{ATE} & \multicolumn{2}{c}{Relative Error} \\
\cline{3-5} \cline{6-7}
& & S1 & S2 & S3 & Trans. (\%) & Rot. (°/m) \\
\hline
\multirow{2}{*}{None}
& SONIC    & 36.6 & 113.3 & 69.8 & 137.65 & 2.45 \\
& ISOPoT   & \textbf{8.8} & \textbf{12.7} & \textbf{16.7} & \textbf{22.76} & \textbf{0.99} \\
\hline
\multirow{4}{*}{\shortstack[l]{Odom.\\+ Mag.}}
& /        & 5.8 & 12.5 & 6.5 & 19.07 & 0.0 \\
& DISO     & 5.3 & 6.1 & 10.9 & 13.90 & 0.44 \\
& SONIC\footnotemark[1] & 7.0 & 11.2 & 13.7 & 22.83 & 0.0 \\
& ISOPoT   & \textbf{3.2} & \textbf{3.5} & \textbf{4.6} & \textbf{9.69} & 0.0 \\
\hline
\end{tabular}
\end{threeparttable}
\caption{ATE and Relative Error on the Aracati 2017 dataset. Results are reported separately for the sonar-only and odometry+magnetometer-assisted settings. The row ``/'' denotes the auxiliary odometry+magnetometer estimate without sonar refinement. RE is evaluated on trajectory segments of length \(10\%\) of the full sequence. Lower is better.}
\label{tab:AracatiCombined}
\end{table}

\subsubsection{Portoroz 2025}

The Portoroz 2025 results, shown in Table~\ref{tab:PortorozCombined}, make the limitations of traditional keypoint-based methods even more apparent. In seabed-dominated scenes with weak local texture, SONIC fails to produce reliable sonar-only odometry. ISOPoT, in contrast, remains stable on both sequences and achieves substantially lower ATE and RE.

\begin{figure}[!t]
   \centering
   \begin{subfigure}[b]{0.88\linewidth}
       \centering
       \includegraphics[width=\linewidth]{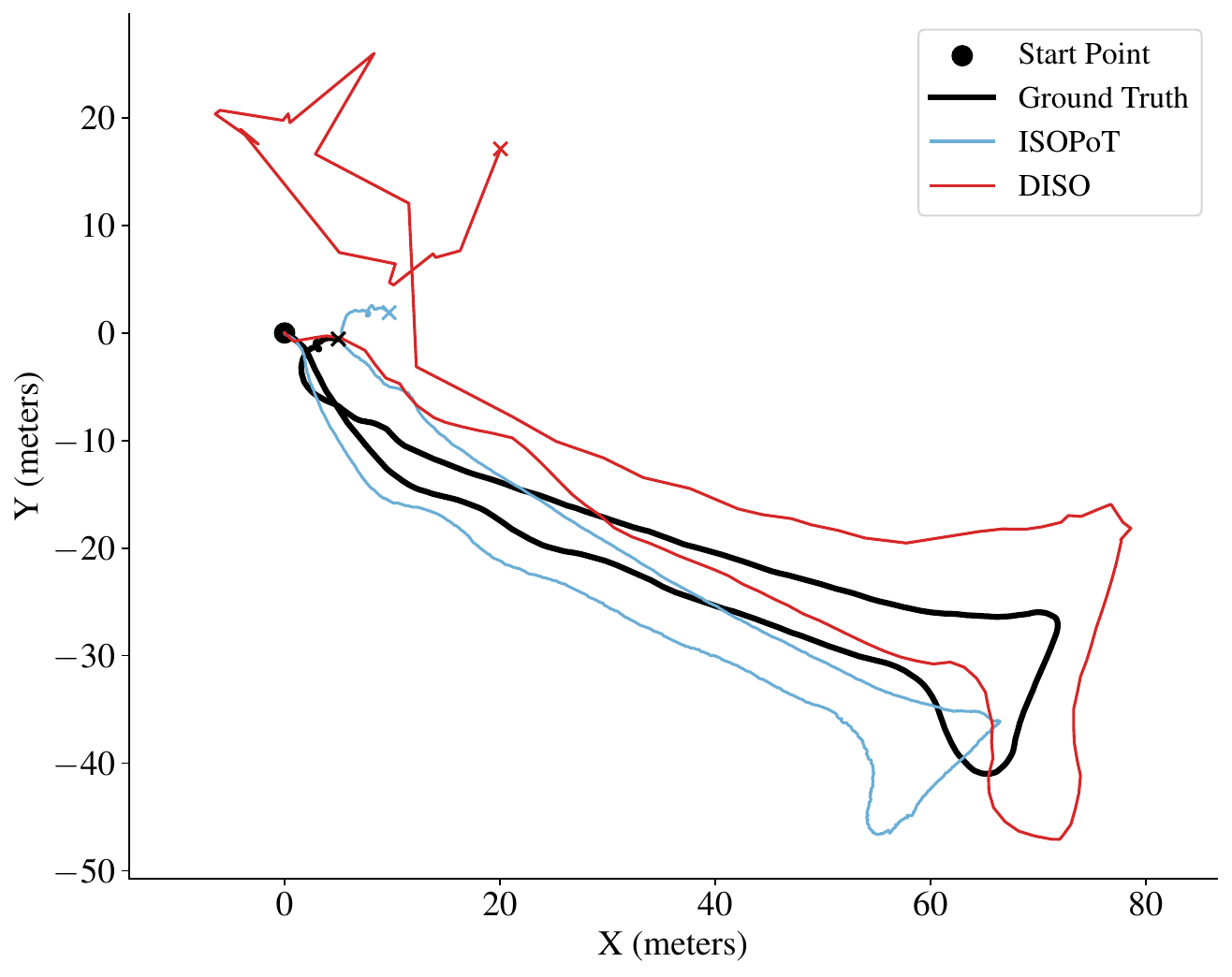}
       \caption{Boot subset}
       \label{fig:Oculus}
   \end{subfigure}
   
   \vspace{0.5em}

   \begin{subfigure}[b]{0.88\linewidth}
       \centering
       \includegraphics[width=\linewidth]{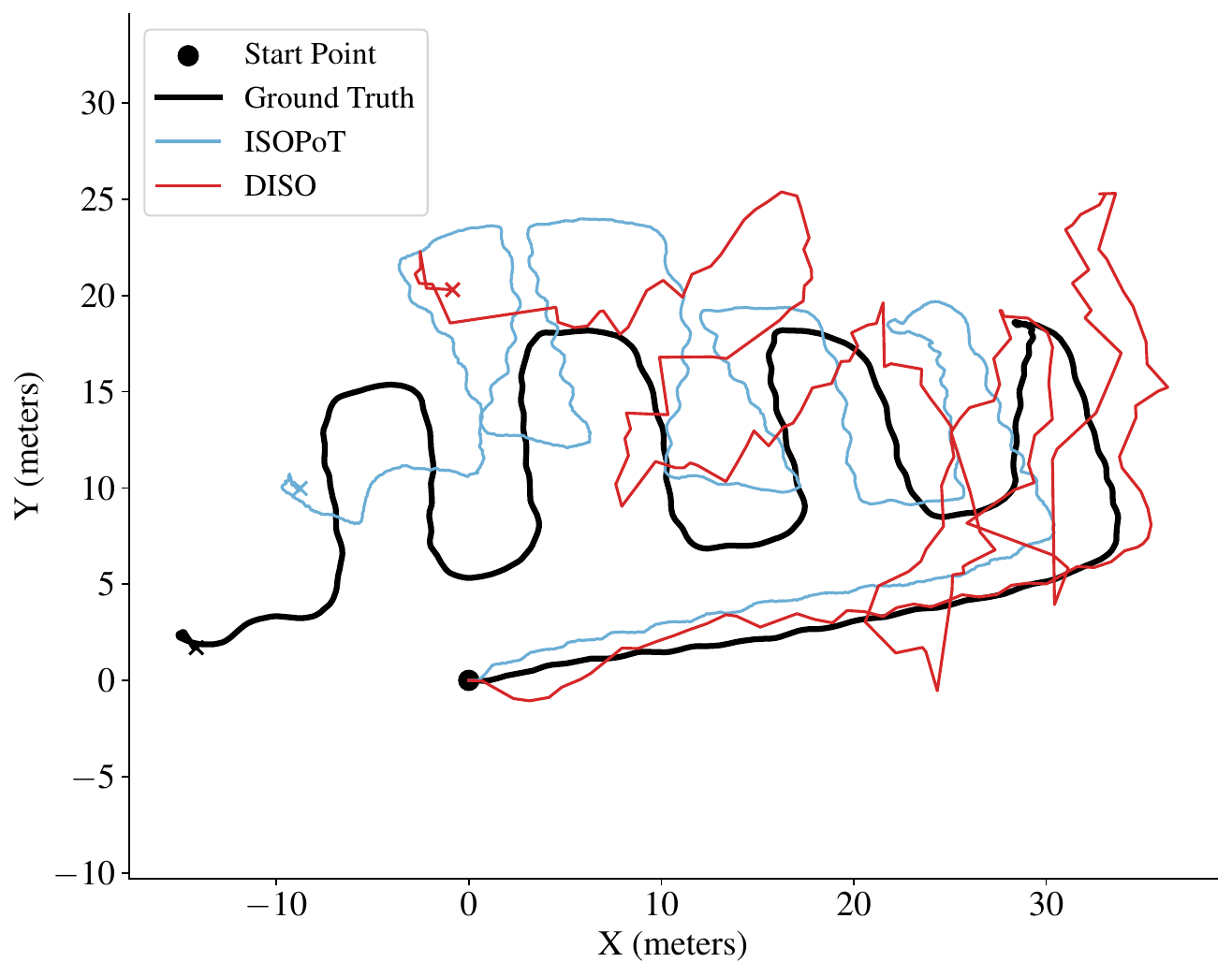}
       \caption{Lawnmower subset}
       \label{fig:lawn2}
   \end{subfigure}
  
   \caption{Qualitative comparison of DISO and ISOPoT on the two Portoroz 2025 sequences. ISOPoT produces smoother and more coherent trajectories, whereas DISO is more strongly affected by failures in image-based refinement.}
   \label{fig:combined}
\end{figure}

DISO performs better than SONIC, primarily because it benefits from external odometry and magnetometer input. However, even in this assisted setting it remains less accurate than ISOPoT on both Portoroz 2025 sequences. This is particularly notable because ISOPoT uses only sonar imagery, whereas DISO has access to auxiliary motion information. The qualitative trajectories in Figure~\ref{fig:combined} are consistent with this observation: DISO often follows the auxiliary estimate but introduces sharp corrections when image-based refinement becomes unreliable, whereas ISOPoT produces smoother and more coherent trajectories.

\begin{table}[!t]
\centering
\small
\setlength{\tabcolsep}{2.5pt}
\begin{threeparttable}
\begin{tabular}{llcccc}
\hline
\multirow{2}{*}{Aux. info.} & \multirow{2}{*}{Method} & \multicolumn{2}{c}{ATE} & \multicolumn{2}{c}{Relative Error} \\
\cline{3-4} \cline{5-6}
& & Boot & Lawn. & Trans. (\%) & Rot. (°/m) \\
\hline
\multirow{2}{*}{None}
& SONIC  & 78.5 & 68.8 & 171.28 & 2.47  \\
& ISOPoT & \textbf{6.5}  & \textbf{6.0}  & \textbf{16.18}  & \textbf{0.61}  \\
\hline
Odom. + Mag. & DISO   & 25.5 & 11.4 & 52.75 &  3.54 \\
\hline
\end{tabular}
\end{threeparttable}
\caption{ATE and Relative Error (RE) on the Portoroz 2025 dataset. Results are reported across the Boot and Lawnmower sequences. ISOPoT achieves substantially lower error than the sonar-only baseline and outperforms DISO despite operating without auxiliary sensors. Lower is better.}
\label{tab:PortorozCombined}
\end{table}

\subsection{Ablation Study}

\subsubsection{Point Tracker Comparison}

To isolate the influence of the tracking backbone, we evaluate several point trackers within the same downstream odometry pipeline. Table~\ref{tab:TrackerCombined} shows that TAPNext provides the best overall performance. In particular, it yields the lowest ATE on the Boot sequence and the lowest relative translation and rotation error.

\begin{figure}[!t]
 \centering
    \begin{subfigure}{\linewidth}
      \centering
      \includegraphics[width=0.97\linewidth]{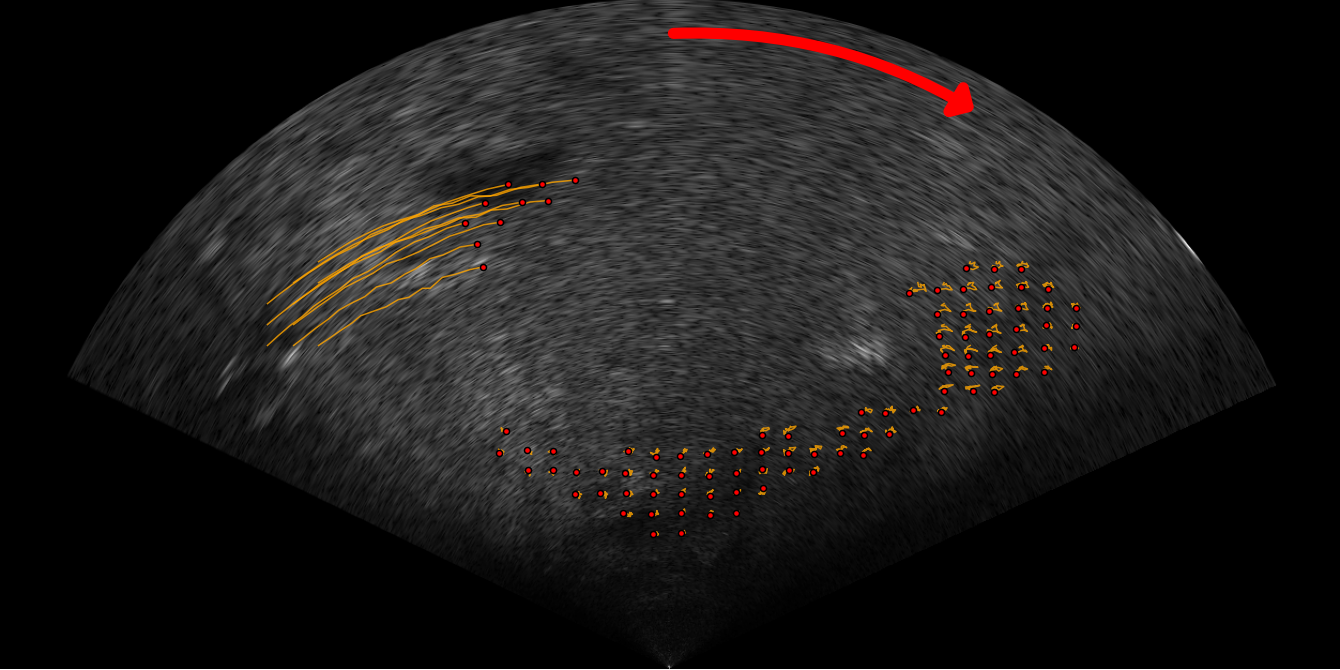}
      \caption{AllTracker}
    \end{subfigure}
    
    \vspace{0.5em}

    \begin{subfigure}{\linewidth}
      \centering
      \includegraphics[width=0.97\linewidth]{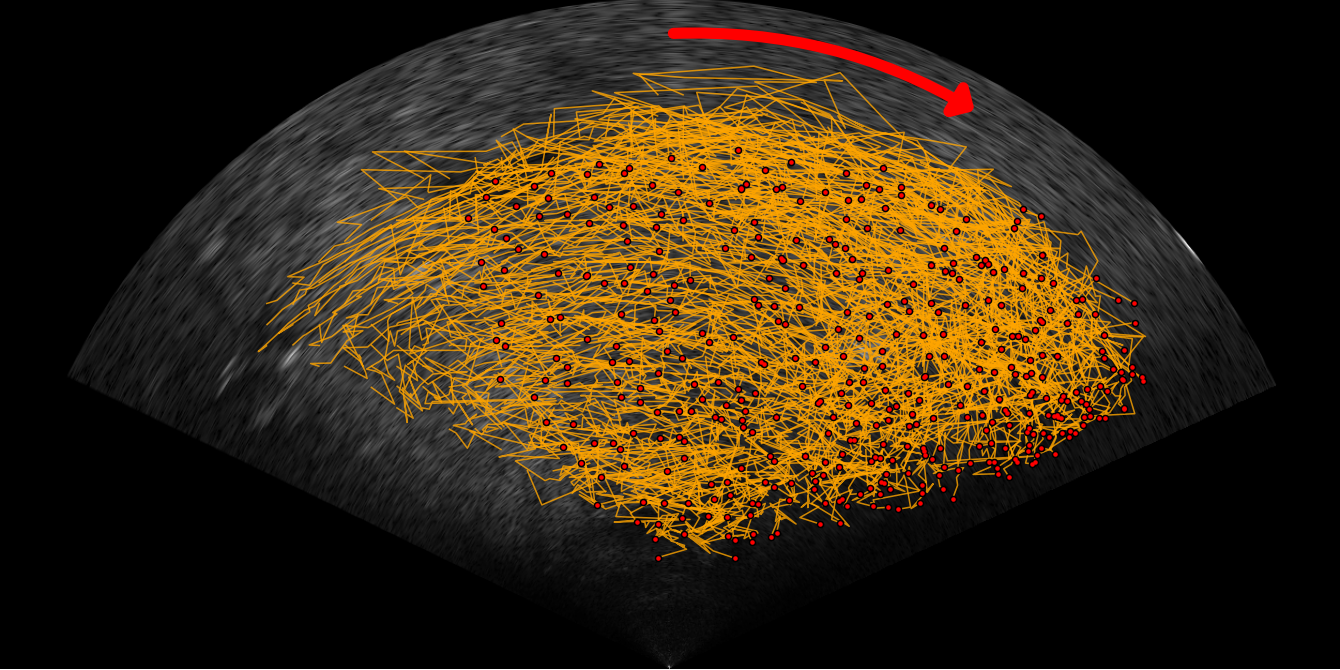}
      \caption{TrackOn2}
    \end{subfigure}
    
    \vspace{0.5em}

    \begin{subfigure}{\linewidth}
      \centering
      \includegraphics[width=0.97\linewidth]{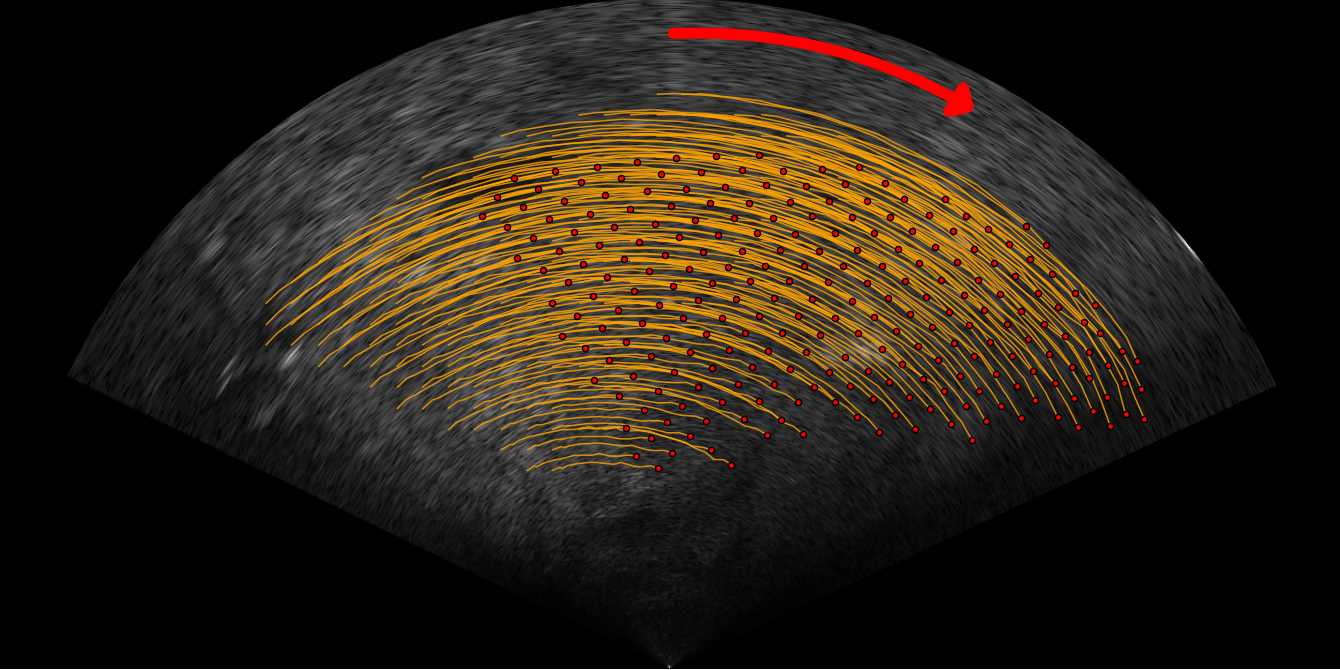}
      \caption{TAPNext}
    \end{subfigure}
 \caption{Qualitative comparison of point trackers on Portoroz 2025 with regular grid initialization. TAPNext predicts the most coherent global motion field. AllTracker captures the correct motion only for a subset of points, while TrackOn2 follows the motion more broadly but with noisier and less smooth trajectories. The red arrow indicates the magnetometer-estimated rotation.}
 \label{fig:TapVsAll}
\end{figure}

The qualitative comparison in Figure~\ref{fig:TapVsAll} explains these differences. AllTracker captures motion only in a limited region of the image, while many points remain nearly stationary or disappear. TrackOn2~\cite{trackon} follows the overall motion more often, but its trajectories are noticeably more jagged and less physically consistent. TAPNext produces the most coherent motion field and is therefore the most suitable tracker for downstream rigid-motion estimation. This supports our choice of TAPNext as the backbone of ISOPoT.

\begin{table}[!t]
\centering
\small
\setlength{\tabcolsep}{6pt}
\begin{threeparttable}
\begin{tabular}{lccc}
\hline
\multirow{2}{*}{Point Tracker} & ATE & \multicolumn{2}{c}{Relative Error} \\
\cline{2-2} \cline{3-4}
& Boot & Trans. (\%) & Rot. (°/m) \\
\hline
AllTracker & 50.3 & 16.87 & 0.84 \\
TrackOn2   & 16.2 & 26.77 & 0.44 \\
TAPNext    & \textbf{6.5}  & \textbf{10.40} & \textbf{0.37} \\
\hline
\end{tabular}
\end{threeparttable}
\caption{ATE and Relative Error (RE) of different point trackers on the Portoroz 2025 Boot sequence using the same downstream odometry pipeline. TAPNext achieves the lowest trajectory error and the most consistent motion estimates. Lower is better.}
\label{tab:TrackerCombined}
\end{table}

\subsubsection{Effect of Individual Subsystems}

We evaluate the contribution of ISOPoT's main components by removing them one at a time: variants without the match refiner, the overlap buffer, and the Grid Point Manager (GPM). Results are reported in Table~\ref{tab:AblationCombined}.

Removing the match refiner produces the largest degradation in ATE, especially in the most difficult part of the Aracati trajectory. This confirms that TAPNext alone is not sufficient: the refinement stage is needed to convert coarse point tracks into a single globally consistent motion estimate. Removing the overlap buffer slightly improves relative translation error, but it worsens ATE, indicating that the buffer mainly improves long-term consistency rather than short-horizon local accuracy. Finally, removing GPM leads to a smaller but consistent degradation, especially in sections with many candidate points, which confirms that spatial coverage and point filtering improve the stability of the RANSAC fit.

Overall, the ablation study shows that the three components address different failure modes and that their combination yields the most robust performance across different operating conditions.

\begin{table}[!t]
\centering
\small
\setlength{\tabcolsep}{2.5pt}
\begin{threeparttable}
\begin{tabular}{lcccccc}
\hline
\multirow{2}{*}{Method} & \multicolumn{4}{c}{ATE} & \multicolumn{2}{c}{Relative Error} \\
\cline{2-5} \cline{6-7}
& S1 & S2 & S3 & Aracati & Trans. (\%) & Rot. (°/m) \\
\hline
ISOPoT      & 3.22 & 3.48 & \textbf{4.57} & \textbf{5.6} & 9.69  & 0.00 \\
No Refiner  & 2.79 & 4.77 & 9.29 & 9.1 & 12.48 & 0.00 \\
No Buffer   & \textbf{2.77} & 5.97 & 5.38 & 10.7 & \textbf{9.20}  & 0.00 \\
No GPM      & 4.43 & \textbf{3.09} & 5.61 & 6.1 & 10.81 & 0.00 \\
\hline
\end{tabular}
\end{threeparttable}
\caption{ATE and Relative Error (RE) of subsystem ablations on Aracati 2017. While the overlap buffer slightly reduces short-range translation error when isolated, the full ISOPoT pipeline achieves the best global trajectory consistency. Lower is better.}
\label{tab:AblationCombined}
\end{table}

\section{CONCLUSIONS}

We introduced ISOPoT, an imaging sonar-based odometry framework that leverages modern point tracking to address the limitations of keypoint-based methods in challenging underwater environments. By combining a state-of-the-art tracker with match refinement, temporal buffering, and structured point management, ISOPoT enables robust motion estimation from noisy, low-semantic sonar imagery. Experiments on the Aracati 2017 and Portoroz 2025 datasets show that ISOPoT consistently outperforms prior approaches, particularly in scenarios with sparse or unreliable keypoints. While auxiliary sensors further improve accuracy, the method achieves strong performance using sonar data alone, demonstrating the effectiveness of point tracking for underwater odometry. Ablation studies highlight the complementary role of each subsystem, and suggest future directions such as incorporating stronger physical constraints and extending the framework toward full SLAM systems.









\end{document}